\documentclass[11pt]{article}

\usepackage[preprint]{acl}

\usepackage{times}
\usepackage{latexsym}

\usepackage[T1]{fontenc}

\usepackage[utf8]{inputenc}

\usepackage{microtype}

\usepackage{inconsolata}

\usepackage{graphicx}
\usepackage{booktabs}
\usepackage{amsmath}
\usepackage{array}
\usepackage{longtable}
\usepackage{tabularx}
\usepackage{xcolor}
\usepackage[most]{tcolorbox}

\newtcolorbox{questionbox}{
  enhanced,
  colback=blue!6,      
  frame hidden,        
  boxrule=0pt,
  sharp corners,       
  left=8pt, right=8pt, top=6pt, bottom=6pt,
  before skip=8pt, after skip=8pt
}
\newcommand{\bench}{\textsc{PayWorthy}}
\newcolumntype{L}[1]{>{\raggedright\arraybackslash}p{#1}}

\title{\bench{}: Can LLMs Predict Which Freelance Jobs an AI Agent Will Complete?}

\author{
  \textbf{Mohammad Reza Rezaei},
  \textbf{Dingxian Wang},
  \textbf{Brett Levert},\\
  \textbf{Celeste Fejzo},
  \textbf{Andrew Rabinovich},
  \textbf{Zhao Chen}
\\
  Upwork Inc.
\\
  \small{\textbf{Correspondence:} \href{mailto:mohammadrezarezaei@upwork.com}{\texttt{mohammadrezarezaei@upwork.com}}}
}

\begin{document}
\maketitle
\begin{abstract}
Before an AI agent takes on a paid job, someone has to judge whether it is likely to succeed. We test whether frontier language models can make this call before any attempt. We propose \bench{} which contains 1{,}494 real freelance job postings, sampled from 18.8M postings made between 2020 and 2026 on a freelancing platform. An agent attempts each job from the posting alone in a single run, and an LLM judge acting as the client decides whether to accept the work. We ask 18 models to forecast how likely the agent is to succeed and measure how well each model's forecasts rank the accepted jobs above the rejected ones (AUC, where 0.5 is a random ranking). The answer depends on how the agent is described. Told what the agent cannot do, every model ranks the jobs well (AUC 0.66--0.86). Told instead that it is a generic agent with a browser and code execution, the same models rank the same jobs about as well as a random ordering (0.45--0.56). These results hold under judges from four other model families (Claude, GLM, Kimi and Qwen), and the informed forecasts also rank the verdicts of two human raters. Code is available at \url{https://github.com/upwork/Payworthy/}.
\end{abstract}

\section{Introduction}
\label{sec:intro}

\begin{figure}[!t]
\centering
\includegraphics[width=\linewidth]{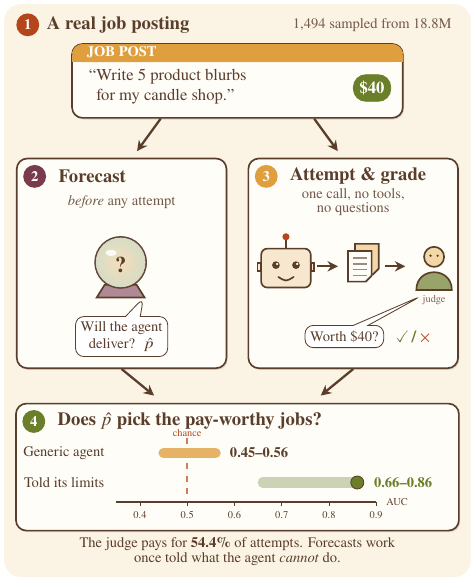}
\caption{\bench{} in brief. (1)~A real job posting. (2)~Before any attempt, a model forecasts the probability $\hat p$ that the agent will succeed. (3)~The agent attempts the job from the posting alone, and an LLM judge acting as the client decides whether to accept the work. (4)~We measure how well $\hat p$ ranks accepted jobs above rejected ones (AUC; 0.5 is a random ranking). When told the agent's limits, the 18 forecasters reach AUC 0.66--0.86; when asked about a generic tool-using agent, they stay near 0.5.}
\label{fig:overview}
\end{figure}

Many agent deployments start with a triage decision:
\begin{questionbox}
\centering\itshape
Which jobs should the agent take on?
\end{questionbox}
A marketplace routing work to agents, a team handing off requests, and an agent bidding for jobs all face the same question before the agent attempts anything:
\begin{questionbox}
\centering\itshape
Will this job end in work the client accepts?
\end{questionbox}
The cheapest way to get this estimate is to ask a frontier model, but we do not know whether such an estimate can be trusted. Existing benchmarks for economically valuable work report how \emph{often} agents succeed on curated tasks \citep{miserendino2025swelancer,patwardhan2026gdpval,mazeika2025rli,yi2025upbench,vidgen2026apexagents,sun2026ale}, not whether anyone can tell in advance \emph{which} tasks they will succeed on. Calibration studies ask whether a model knows its own answer is right after it has answered \citep{kadavath2022language,xiong2024can}; \citet{barkan2025capable} ask before the attempt, but only for coding tasks. No study tests whether a model can forecast, before any attempt, which real, paid jobs an agent will complete to the client's satisfaction.

We built \bench{} to test this (Figure~\ref{fig:overview}). \bench{} contains 1{,}494 job postings sampled from 18.8M real freelance job ads posted between 2020 and 2026. An agent attempts each job, and an LLM judge acting as the client decides whether to accept the result based on the job's requirements. To match the information available to the forecaster, the agent receives only the posting and works in a single response without tools (\S\ref{sec:attempt}). Before any attempt, 18 frontier models forecast how likely the agent is to succeed on each job, and we measure how well each model's forecasts rank the jobs the judge accepts above those it rejects.

How often a judge accepts an outcome depends on the judge as well as on the job and the agent: judges from five model families accept 33--58\% of the same deliverables. We therefore evaluate how well forecasts \emph{rank} jobs, and check that these rankings hold across different judges, with and without repeated attempts, and against two human raters (\S\S\ref{sec:noise}--\ref{sec:robust}).

\begin{figure}[t]
\centering
\includegraphics[width=\linewidth]{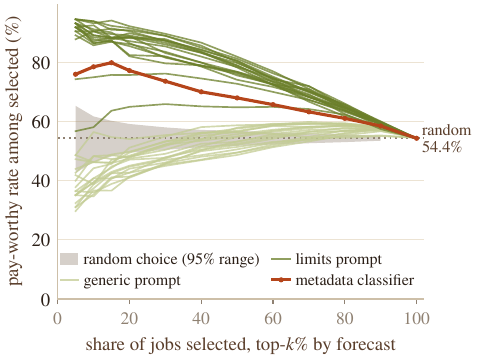}
\caption{Choosing jobs by forecast: the acceptance rate among the top $k\%$ of jobs, ranked by each of the 18 forecasters. The band covers 95\% of 2{,}000 random selections of the same size. At $k=20\%$, limits forecasts reach 65--92\%, generic forecasts 41--54\% (below the band for 15 of 18), and a cross-fitted classifier on posting metadata 77\% (\S\ref{sec:selection}).}
\label{fig:teaser}
\end{figure}

How well models predict this outcome depends on what they are told about the agent. We prompt each forecaster in two ways. The \emph{limits} prompt lists what our agent cannot do, such as receive files from the client or use tools. Given this prompt, all 18 forecasters rank the jobs well (AUC 0.66--0.86), and the fifth of jobs each one rates highest is accepted 65--92\% of the time, against 54\% overall (Figure~\ref{fig:teaser}). The \emph{generic} prompt describes a capable agent with a browser and code execution, as a user might describe an agent without knowing its setup. Given this prompt, the same forecasters rank the jobs about as well as a random ordering (AUC 0.45--0.56), and for 15 of the 18, the fifth of jobs they rate highest is accepted significantly \emph{less} often than a random fifth. Above all, they miss that a job requiring the client's files will fail for an agent that never receives them (\S\ref{sec:why}).

To test if the generic prompt describes a different agent, we also run that agent, with web search, code execution, and a simulated client it can ask for information. The judge accepts its work more often (67\% against 52\% on the same 500 jobs), yet generic forecasts still rank its jobs near random (Appendix~\ref{sec:nearchance}).

\paragraph{Contributions.} (1)~\bench{}: real freelance jobs from 12 categories, sampled by a fixed rule, attempted by an agent, and graded by a simulated client whose verdicts we validate in several ways (\S\S\ref{sec:bench}--\ref{sec:where}). (2)~Evidence that forecast quality depends on how the agent is described, with two of six stated limits accounting for most of the gain (\S\S\ref{sec:harness}--\ref{sec:ablation}). (3)~Evidence that, under the generic description, forecasters do not account for an agent that never receives the client's files, a failure that larger models, reasoning models, and averaging do not fix (\S\ref{sec:why}, Appendix~\ref{sec:nearchance}).
\section{Related Work}
\label{sec:related}

\paragraph{Benchmarks of agentic and economically valuable work.}
Agent benchmarks such as SWE-bench \citep{jimenez2024swebench}, WebArena \citep{zhou2024webarena}, GAIA \citep{mialon2023gaia}, $\tau$-bench \citep{yao2024taubench}, OSWorld \citep{xie2024osworld}, and TheAgentCompany \citep{xu2024theagentcompany} use tasks whose success a program can check. Recent benchmarks grade economically valuable or freelance work against payouts or rubrics: SWE-Lancer \citep{miserendino2025swelancer}, GDPval \citep{patwardhan2026gdpval}, the Remote Labor Index \citep{mazeika2025rli}, UpBench \citep{yi2025upbench}, APEX-Agents \citep{vidgen2026apexagents}, and Agents' Last Exam \citep{sun2026ale}. All report how often agents succeed on tasks that people chose or wrote (Table~\ref{tab:benchcmp}). We instead sample jobs by a fixed rule (\S\ref{sec:samples}) and ask whether a model can tell in advance which of them an agent will succeed on. Studies of occupational exposure and task duration \citep{eloundou2024,felten2021,handa2025economic,tomlinson2025working,kwa2025measuring} and of freelance platforms \citep{hui2024,demirci2025,teutloff2025} quantify what AI could do or is already doing; we draw on the same kind of platform data, but as a source of jobs for agents to attempt.

\begin{table*}[t]
\centering\small
\caption{\bench{} among benchmarks of agentic and economically valuable work. ``Grader'': what decides success. ``Prices'': tasks carry real posted prices. ``Sampled'': items are drawn at random or by stratified sampling from posted work, with no person choosing or screening them (\bench{} applies one fixed automatic filter; \S\ref{sec:samples}). ``Forecast'': models predict success before the attempt.}
\label{tab:benchcmp}
\setlength{\tabcolsep}{5pt}
\begin{tabular}{@{}lrllccc@{}}
\toprule
Benchmark & Tasks & Domain & Grader & Prices & Sampled & Forecast \\
\midrule
SWE-bench~\citep{jimenez2024swebench}         & 2{,}294 & Python repos          & unit tests        & no  & no  & no  \\
$\tau$-bench~\citep{yao2024taubench}          & 165     & retail, airline       & DB state + rules  & no  & no  & no  \\
SWE-Lancer~\citep{miserendino2025swelancer}   & 1{,}488 & software              & E2E tests         & yes & no  & no  \\
GDPval~\citep{patwardhan2026gdpval}           & 1{,}320 & 44 occupations        & expert comparison & no  & no  & no  \\
RLI~\citep{mazeika2025rli}                    & 240     & remote freelance      & expert review     & yes & no  & no  \\
UpBench~\citep{yi2025upbench}                 & 322     & freelance jobs        & expert rubrics    & yes & no  & no  \\
APEX-Agents~\citep{vidgen2026apexagents}      & 480     & professional services & expert rubrics    & no  & no  & no  \\
\midrule
\bench{} (ours)                               & 1{,}494 & 12 categories         & simulated client  & yes & yes & yes \\
\bottomrule
\end{tabular}
\end{table*}

\paragraph{Predicting success in advance.}
Calibration work asks whether a model knows when it is right by scoring the confidence it reports after answering, whether stated in words \citep{lin2022teaching,xiong2024can} or obtained by self-evaluation \citep{kadavath2022language,tian2023just}. Related work tests whether models recognize questions they cannot answer \citep{yin2023large}, lets a model abstain when it is likely to be wrong \citep{kamath2020selective}, and selects items while bounding the share of failures among them \citep{jin2023selection}; our top-$k$ selection (\S\ref{sec:selection}) extends this last line of work. Closest to us, \citet{barkan2025capable} ask models before the attempt whether they will solve coding tasks, and find weak discrimination. Our forecasters instead estimate how likely the agent described in the prompt is to succeed on paid jobs across 12 top-level categories (Table~\ref{tab:percat}), well beyond coding.

\section{The \bench{} Benchmark}
\label{sec:bench}

\subsection{Jobs}
\label{sec:samples}
\begin{figure}[t]
\centering
\includegraphics[width=\linewidth]{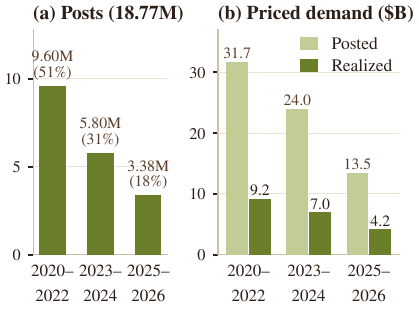}
\caption{Corpus and priced demand. (a)~Posts by era (18.8M in total). (b)~Priced demand in billions of dollars (\$69.1B posted, \$20.4B realized in total).
Realized demand counts only postings with at least one awarded contract.}
\label{fig:corpus}
\end{figure}

The jobs are drawn from \textbf{anonymized} daily extracts of a large freelance marketplace spanning more than five years, from 2020 through mid-2026: 18.8M postings requesting \$69.1B worth of work (Figure~\ref{fig:corpus}). Each posting has a title, a description, skill tags, a fixed budget or an hourly rate range, an expected duration, a required expertise level and a three-level category label (12 top-level categories). For an hourly posting, the contract value is calculated as the midpoint rate times the hours implied by the expected duration (Appendix~\ref{app:corpus}).

We draw the sample in three steps (Table~\ref{tab:funnel}). First, to keep the candidate pool small and varied, we take the postings of the 1st and the 15th of each month and keep at most the first 15 postings per fine-grained category and day, which limits how much the most common categories contribute. Second, we draw 4{,}000 postings from this pool, stratified by top-level category and month. Third, we drop descriptions under 150 characters, shuffle the rest and go through them in order, keeping a posting only if its job has a \emph{defined deliverable}, a product the agent could hand over rather than an ongoing service, as judged by at least four of the seven base forecasters (\S\ref{sec:forecasts}). We stop at 1{,}500 kept postings; the agent returned a deliverable for 1{,}494. Because of the cap, less common categories are over-represented; reweighting to the corpus shares of postings or dollars puts the acceptance rate at 51--58\%, against 54\% as sampled (Table~\ref{tab:selection}). Table~\ref{tab:percat} gives the jobs and acceptance rate per category.

\subsection{Attempt (Task A)}
\label{sec:attempt}
For Task A, the agent is based on claude-opus-5 and sees the posting as a freelancer would and is asked to take the job from acceptance to delivery while working in a single response without tools (a long response is continued across several calls; see Limitations); it can hand over only files it writes as text or code (documents, spreadsheets and images are built from that output). Additionally, it must not claim work it has not done: anything that needs access, files or a medium it lacks goes into a note to the client. This constraint of working from the posting alone provides the proper information the forecaster has. Without a live client or access to real accounts, each attempt is cheap, safe to run on real postings and easy to repeat (\S\ref{sec:noise}). The cost is that jobs which need the client's material are hard for this agent, so Appendix~\ref{sec:nearchance} also runs an agent that has tools and can ask a simulated client.

\subsection{Grading}
\label{sec:grading}
An LLM judge, gpt-5, sees the posting with its budget and the whole deliverable, and answers four yes-or-no questions: is there a deliverable at all (\textbf{produced}); does it address what the posting asked for (\textbf{on-spec}); would a reasonable client accept it without redoing it (\textbf{client-plausible}); and would the client pay the posted budget for it without renegotiating (\textbf{pay-worthy}; yes if payment is more likely than not). It also flags as \textbf{fabricated} any deliverable that claims work the agent could not have done, such as testing the client's systems; such a deliverable is never pay-worthy. The judge votes three times, and the majority pay-worthy vote is the label; a job is \emph{accepted} when it is yes. The pay-worthy and client-plausible verdicts agree on 98\% of jobs, so the budget question adds almost nothing to whether a reasonable client would accept the work. With pay type controlled for, larger budgets are accepted less often (\S\ref{sec:bycat}), but larger jobs may simply be harder; separating the two would require grading the same work under different budgets.

Each vote that rejects a job also names one reason from a fixed list, and a rejected job takes the most common one; four rejected jobs are left unlabeled (Figure~\ref{fig:outcomes}b). We group the reasons into six causes (Appendix~\ref{app:impl}): scoping, missing information (the job needs files, access or preferences that the posting does not supply), modality or harness, verification, execution and fabrication. Only four deliverables (0.3\%) are flagged as fabricated; dropping them changes no forecasting result (Appendix~\ref{app:extra}).

\textbf{Metrics.} Our main measure is AUC: the probability that a forecaster rates a randomly chosen accepted job above a randomly chosen rejected one. A forecast unrelated to the outcome scores 0.5 whatever the acceptance rate and however high its forecasts run, so we call 0.5 a random ranking. When forecasts choose jobs, we report the acceptance rate among the top $k\%$; there the reference is the overall rate, 54.4\%, and the range random choices produce. We also report calibration error (10-bin ECE) and the mean forecast divided by the acceptance rate. Confidence intervals are 95\% percentile bootstrap intervals over jobs (2{,}000 resamples, paired for differences).

\subsection{Judge agreement}
\label{sec:reliability}
The judge's three votes agree closely (Fleiss $\kappa=0.86$ on pay-worthy). To check that the verdict is not one model's taste, judges from four other model families (glm-5.3, claude-sonnet-5, qwen3.8-max and kimi-k3) grade the same deliverables with the same prompt. Their acceptance rates differ, but all five reach the same verdict on 65\% of jobs, and their majority agrees with gpt-5 at $\kappa=0.83$ (Figure~\ref{fig:judges}). Section~\ref{sec:noise} checks that the verdict follows the work, \S\ref{sec:robust} repeats the main results under each judge, and Appendix~\ref{sec:hapi} compares the judge with real clients.

\subsection{Forecasts (Task B)}
\label{sec:forecasts}
Before any attempt, a forecaster sees the posting and gives $\hat p$, the probability that the agent will complete the job to a standard the client would accept and pay for, with five annotations of the job (Table~\ref{tab:axes}). It is told about the agent in one of two ways (\S\ref{sec:harness}) and never sees the judge.

We query 18 forecasters, chosen to test whether model size, reasoning or model family changes the result. Seven \emph{base} forecasters of different sizes and kinds come from two providers (gpt-5, gpt-5-mini, gpt-5-nano, gpt-4.1, gpt-4o, o3 and claude-sonnet-4.5); their annotations also define the sample. Eight more are mostly newer and larger OpenAI and Anthropic models, including claude-opus-5, which thus forecasts its own attempts. The last three, glm-5.3, kimi-k3 and qwen3.8-max, come from the families of the other judges, so we can test whether forecasters favor the judge from their own family (\S\ref{sec:robust}). Appendix~\ref{app:mech} gives the prompts, Table~\ref{tab:models} the models and Table~\ref{tab:settings} every setting.

\section{Which Jobs the Agent Completes}
\label{sec:where}

\begin{figure}[t]
\centering
\includegraphics[width=\linewidth]{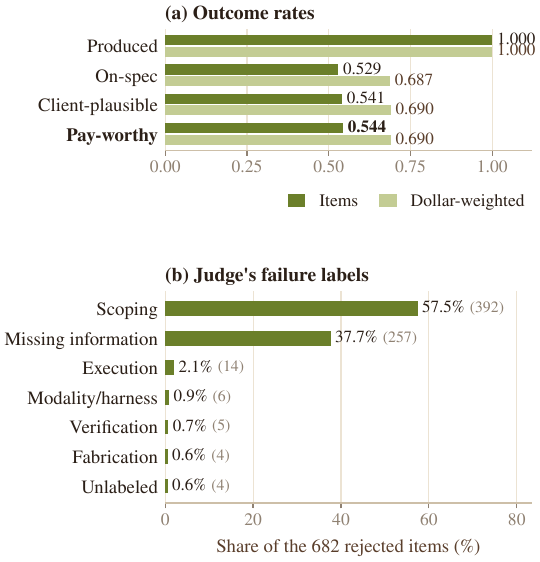}
\caption{Agent outcomes on \bench{} (solver claude-opus-5; judge gpt-5 reading in full, three votes per job). (a)~Share of jobs meeting each of the judge's criteria (dark) and the same share weighted by contract value over priced jobs (light; mostly reflects pay type, Appendix~\ref{app:corpus}). Each criterion is a separate majority vote, so the shares need not fall from one criterion to the next. (b)~The judge's failure reasons for the 682 rejected jobs (counts in parentheses).}
\label{fig:outcomes}
\end{figure}

The agent hands over files for every job and the judge accepts 54.4\% of them (Figure~\ref{fig:outcomes}a). Most rejections are for scoping (57\%) or missing information (38\%; Figure~\ref{fig:outcomes}b). The judge picks these reasons from names without definitions, so we use them only in aggregate.

\begin{figure*}[t]
\centering
\includegraphics[width=\textwidth]{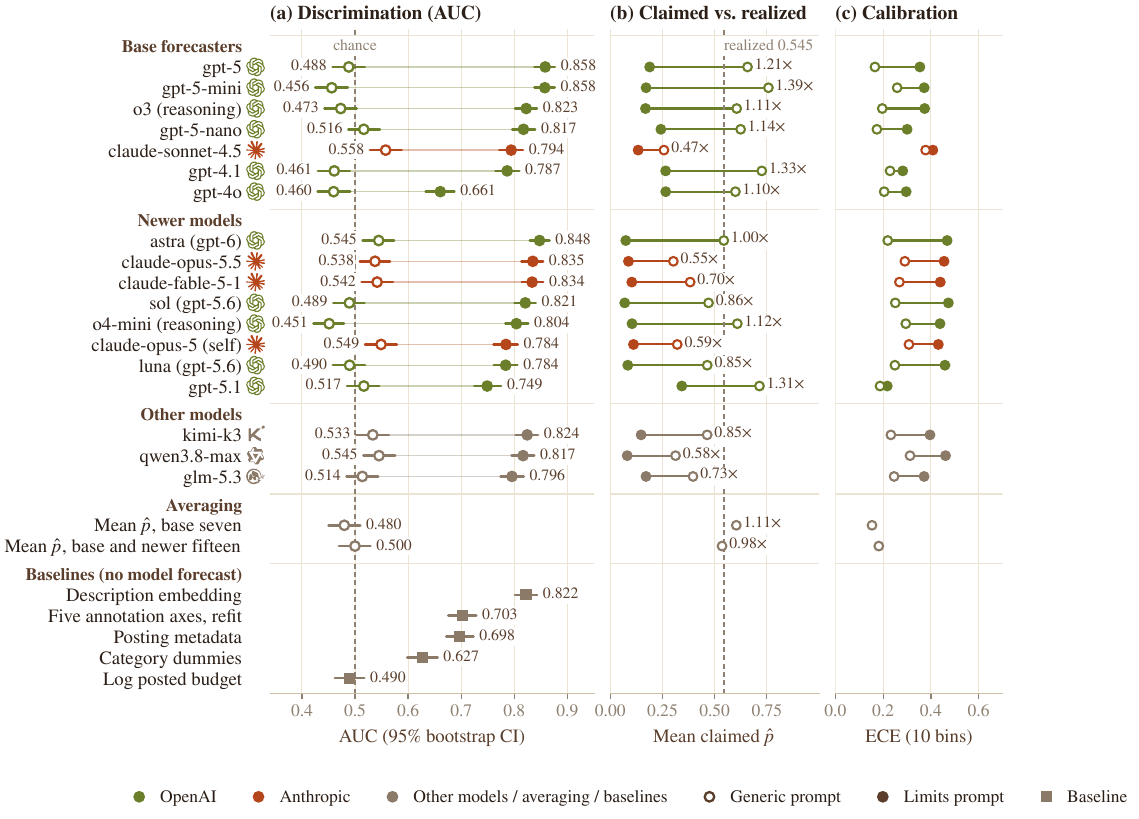}
\caption{Can models tell in advance which jobs the judge will accept? Each forecaster rates every posting with the generic prompt (open circles) and the limits prompt (filled circles), compared with the judge's verdict on the attempt (headline label); forecasters are sorted by limits AUC within each group.
(a)~AUC with 95\% bootstrap CI; 0.5 is a random ranking. Every
forecaster's gain from the limits prompt excludes 0 (paired bootstrap). (b)~Mean forecast against the acceptance rate (dashed line, 0.545 on these postings); the multiplier is the mean generic forecast divided by that rate, so values above 1 are overclaims.
(c)~10-bin expected calibration error. Averaging rows average the generic forecasts of the seven base forecasters or of the
fifteen OpenAI and Anthropic forecasters. Baselines use no forecast and are 5-fold cross-validated, except log posted budget, which ranks the 350 hourly jobs without a stated rate as the cheapest
(among priced jobs it reaches 0.58); description embedding is a logistic regression on text embeddings of the posting.}
\label{fig:forecasters}
\end{figure*}

\subsection{By category, pay type, budget and need}
\label{sec:bycat}\label{sec:budget}
Acceptance varies widely by category, from 12\% in Translation and 23\% in Admin Support to 74\% in IT \& Networking (Table~\ref{tab:percat}), yet category alone ranks jobs only modestly (Figure~\ref{fig:forecasters}). Hourly jobs are accepted more often than fixed-price ones (66\% against 43\%). Among fixed-price jobs, whose budgets the clients set, larger budgets are not accepted more often, and in a regression that also controls for pay type and for five annotations of each job, a larger budget lowers the odds of acceptance (Table~\ref{tab:mech}). Description length and skill count have no significant effect: a longer brief is not a longer task (cf.\ \citealp{kwa2025measuring}).

\begin{table}[t]
\centering\small
\caption{Jobs per top-level category in \bench{}, their share of the 1{,}494 jobs and the share the judge accepts (headline label), both in percent, with the same split by pay type.}
\label{tab:percat}
\setlength{\tabcolsep}{4pt}
\begin{tabular}{@{}lrrr@{}}
\toprule
Category & Jobs & Share & Accepted \\
\midrule
Design \& Creative & 385 & 25.8 & 42.9 \\
Web, Mobile \& Software Dev & 248 & 16.6 & 66.9 \\
Engineering \& Architecture & 204 & 13.7 & 55.9 \\
Writing & 136 & 9.1 & 55.9 \\
Data Science \& Analytics & 109 & 7.3 & 67.9 \\
Accounting \& Consulting & 84 & 5.6 & 65.5 \\
Sales \& Marketing & 80 & 5.4 & 62.5 \\
IT \& Networking & 74 & 5.0 & 74.3 \\
Admin Support & 70 & 4.7 & 22.9 \\
Legal & 51 & 3.4 & 56.9 \\
Translation & 41 & 2.7 & 12.2 \\
Customer Service & 12 & 0.8 & 58.3 \\
\midrule
Fixed-price & 754 & 50.5 & 43.2 \\
Hourly & 740 & 49.5 & 65.7 \\
\midrule
Total & 1{,}494 & 100.0 & 54.4 \\
\bottomrule
\end{tabular}
\end{table}

What the job needs from the client matters more. An LLM annotation of each posting (the \emph{needs label}) sorts jobs into four kinds: \emph{self-contained} jobs, which need nothing beyond the posting, and jobs that need the client's files, access to the client's systems, or a medium the agent cannot produce. The judge accepts 82\% of self-contained jobs, against 42\% of jobs that need the client's files and 32\% of jobs that need another medium (Table~\ref{tab:needs}). Like the agent, the judge never sees these files, so the 42\% likely overstates success: human raters are far stricter on such jobs (\S\ref{sec:why}).

\subsection{Short briefs leave the agent short of information}
\label{sec:missinginfo}
An agent working from the posting alone has only what the brief supplies. Retrieval can add public evidence \citep{vendi-rag,amg-rag}, but not the client's own files or preferences, so a job whose brief leaves them out can fail for lack of information. Rejections for missing information hit the shortest third of briefs about 40\% more often than the longest third (Table~\ref{tab:needs}). For our agent, a short brief is a risk, not a sign of an easy job. Human raters show the same pattern more strongly (\S\ref{sec:robust}).
\section{Can Models Tell in Advance?}

\label{sec:harness}
Each forecaster rates every job twice, under two descriptions of the agent (both prompts are in Appendix~\ref{app:mech}). The \emph{generic} prompt describes an agent ``equipped with a browser, code execution, and file read/write, that could take this job from acceptance to delivery with no human in the loop besides the client''. The \emph{limits} prompt describes the agent we actually run, through six limits: it ``works from the posting alone'' (L1); it ``cannot ask the client clarifying questions or get feedback'' (L2); it has ``no access to the client's accounts, websites, systems, data or physical devices'' (L3); it ``can produce only text-encodable output'' (L4); it ``submits its work as named files'' (L5); and it ``works in a single response with no tools'' (L6).

Told the agent's limits, all 18 forecasters rank the jobs well (AUC 0.66--0.86; Figure~\ref{fig:forecasters}a); under the generic prompt, they rank them about as well as a random ordering (0.45--0.56). Every forecaster gains from the limits prompt, by 0.30 AUC on average.

Limits forecasts also hold up against classifiers trained on outcomes. The best such classifier, a logistic regression on text embeddings of the posting, reaches AUC 0.82 after training on about 1{,}200 labeled jobs. The three best limits forecasts beat it without any labeled jobs. Adding any limits forecast to the classifier improves it further, while adding a generic forecast does not (Table~\ref{tab:limits}).

Limits forecasts rank jobs well but run low: their means range from 0.07 to 0.34, against an acceptance rate of 54\%. The forecasters never see the judge, so this reflects their expectations of the agent; recalibrating on 100 labeled jobs removes most of the gap (Table~\ref{tab:protocol}).

\subsection{Which limits matter}
\label{sec:ablation}
For four forecasters, we reran the limits prompt with each limit removed, with each limit alone, and with a placebo that replaces the limits by six neutral statements of similar length (Figure~\ref{fig:ablation}). Two limits carry most of the gain. Stating only L6 (single response, no tools) or only L1 (posting only) recovers most of it, and removing either of them, or L5 (named files), lowers the AUC of all four forecasters. The placebo does no better than the generic prompt, so the gain comes from what the limits say, not from the form of the prompt.

\paragraph{More than spotting missing files?} For an agent that receives nothing but the posting, a job that needs the client's files is likely to fail, and a fixed rule that ranks self-contained jobs above all others, using no labeled jobs, reaches AUC 0.64. Limits forecasts do more than this. Seventeen of the 18 beat the rule, and all 18 rank jobs significantly better than random within groups of jobs that share a needs label, where the rule cannot tell jobs apart. Every forecaster also gains significantly from the limits prompt on the self-contained jobs, where nothing is missing, and when rejections for missing information are set aside (Table~\ref{tab:limits}).

\subsection{Forecast level, averaging and job selection}
\label{sec:prior}\label{sec:selection}
Forecasters from different model families give very different numbers for the same jobs, but differ far less in how well they rank them (Figure~\ref{fig:forecasters}). Under the generic prompt, most OpenAI forecasters overclaim and the others underclaim, yet all rank near random; under the limits prompt, all run low, yet all rank well. A threshold for taking on jobs must therefore be calibrated for each forecasting model (\S\ref{sec:harness}). Calibration cannot change AUC, so it would not rescue generic forecasts; nor does averaging them, because they agree with one another more than with the outcome.

The same holds when forecasts are used to choose jobs (Figure~\ref{fig:teaser}). Among the fifth of jobs each forecaster rates highest, limits forecasts find 65--92\% accepted, against 54\% for a random choice, and 16 of the 18 also beat a classifier trained on posting metadata. Generic forecasts find 41--54\%, which for 15 of the 18 is below 97.5\% of random choices of the same size. Their top fifth over-represents jobs that need the client's files (58\% of picks against 43\% of all jobs; 31\% accepted), and the same forecaster's limits forecasts put 37\% of these picks in their bottom half (medians; Table~\ref{tab:track}).

\subsection{How noisy is the label, and does the judge read the work?}
\label{sec:noise}
On 400 jobs, the agent made two more attempts and the judge voted again on each deliverable (Appendix~\ref{app:noise}). Attempts at the same job agree at Cohen $\kappa=0.81$ and fresh votes at $\kappa=0.87$, so the verdict mostly reflects the job and not the attempt or the vote. The agent's other two attempts predict the verdict on a third at AUC 0.94, a practical ceiling for any forecast made in advance; the best forecast (0.86) gets most of the way there.

The verdict also follows the work. Keeping each posting but swapping in a weaker model's attempt, the first quarter of the agent's own deliverable, or another job's deliverable cuts acceptance from 54--57\% for the agent's work to 1--19\%.

\subsection{Robustness}
\label{sec:robust}
\paragraph{Other judges and solvers.} Under seven other labels of the same deliverables (gpt-5 reading only the opening, each other judge, the five judges' majority, and the 978 jobs all five agree on), generic forecasts stay near random and every forecaster gains significantly from the limits prompt (Table~\ref{tab:labels}, Figure~\ref{fig:judges}). The same holds for two other solvers, claude-fable-5-1 and the weaker claude-haiku-4.5, on the 400 jobs of \S\ref{sec:noise} (Appendix~\ref{app:judges}).

\paragraph{Real clients.} On 322 jobs outside the benchmark where a real client paid for a freelancer's work, the judge accepts only 16.5\% of that paid work, but 27.8\% of claude-opus-5's attempts at the same jobs, and it does not reliably tell paid from rejected submissions (Appendix~\ref{sec:hapi}). We leave open whether the agent's work is better, as the judge sees only the final submission, not the exchanges and files behind it; the anchor shows the judge's bar is not a real client's, hence we evaluate rankings.

\paragraph{Human raters.} Two raters with experience in the jobs' categories scored 149 jobs the judge had accepted, using the judge's criteria and without seeing its verdict (Appendix~\ref{app:human}). They accept only 35\% of this work, and three quarters of their rejections cite missing information, more often for short briefs (58\% of jobs in the shortest third against 36\% in the longest). Within this set, which the judge does not separate, limits forecasts rank the raters' verdicts better than generic forecasts (mean AUC 0.65 against 0.54), and every forecaster gains. Since the judge accepted all these jobs, a classifier trained on its verdicts ranks the raters' verdicts near random (0.53); limits forecasts thus capture what raters reward beyond the judge, mainly the need for client material.

\paragraph{Biases shared with the judge.} Forecasters rank their own family's judge better by less than 0.01 AUC, far less than the limits gain. The judge prefers longer deliverables \citep{dubois2024length}, but limits forecasts keep their AUC within each fifth of deliverable length. The results also hold on jobs whose success is easy to check, without fabrication-flagged deliverables, and on our paraphrased postings (Appendix~\ref{app:judges}, Table~\ref{tab:selection}).

\section{Why Generic Forecasts Fail}
\label{sec:why}

Human raters widen the gap of \S\ref{sec:bycat} between self-contained jobs and those that need the client's files, accepting 53\% against 17\% of the judge's accepts (Appendix~\ref{app:human}). Generic forecasts do not rate self-contained jobs above those that need the client's files (AUC 0.41--0.57), while limits forecasts do (0.69--0.87; Table~\ref{tab:track}). Generic forecasters do read the posting, since they rate low the jobs that need a medium the agent cannot produce; what they do not know is that the agent will not receive the client's material. Job size (budget, description length, skill count, pay type, expertise level) explains at most 3\% of their variance. Instead, a generic forecast follows the forecaster's own annotations of the job \citep{mcinerney2023chill,maricl}, above all how much account, payment or physical access it thinks the job needs. In their one-line reasons, generic forecasters often name missing client material as a risk without lowering their forecast; limits forecasters lower it (Table~\ref{tab:track}).

A forecaster picturing an agent that can ask for files has reason to discount missing ones; for an agent with only the posting, they are decisive (\S\ref{sec:missinginfo}). But the mismatch does not explain the whole failure. Setting aside rejections for missing information raises generic AUCs only a little (Table~\ref{tab:needs}; a rough check, since the judge assigns that reason after seeing the work), and generic forecasts are near random for the tool agent too, whose client can supply missing material (Appendix~\ref{sec:nearchance}). Generic forecasts thus fail on two counts: they assume the agent gets the client's material, and they leave unused signal in their own annotations (Appendix~\ref{app:extra}).

\section{Conclusion}
Whether frontier models can predict which freelance jobs a client-simulating judge will accept from an agent depends on how the agent is described. Told what the agent cannot do, all 18 forecasters rank the jobs well. Asked about a generic tool-using agent, they rank them no better than random, missing that the agent never receives the client's files; larger models, reasoning models and averaging do not help. The best forecasts still trail repeated attempts (AUC 0.86 against 0.94). A model used for triage should be told the agent's limits. Our results do not measure real payment or whether prices are fair.

\section*{Limitations}
\label{sec:limits}

\textbf{A judge's verdict.} The label is an LLM judge's acceptance of the work, not a real client's payment, and it does not separate acceptance from price (\S\ref{sec:grading}). Like the agent, the judge never sees the client's files, so it may accept work that would not fit them; human raters accept only about a third of the work the judge accepts, and far less on jobs that need those files.
\textbf{Context size.} Long deliverables are written across several calls and may differ from ones written in a single call, and nine (0.6\%) are still cut at 64{,}000 tokens.
\textbf{Models change.} Most forecasters and judges are called under provider aliases that can be updated or retired (Table~\ref{tab:models}), so a re-run may not reproduce our numbers exactly; the released forecasts and votes fix the ones we report.
\textbf{Human raters.} Two raters scored 149 of the judge's accepts, each covering six categories, so we cannot measure agreement between raters, and they did not rate the jobs the judge rejected.

\section*{Ethical Considerations}
\textbf{Data.} Postings come from a commercial freelance platform. Our access, including the real submissions of the anchor
(Appendix~\ref{sec:hapi}), is covered by an internal agreement that
permits research use and the release of paraphrased surrogates, and the original files are used under license; the raw extract is not
redistributed. Released postings are paraphrased surrogates, and the platform name and client-identifying details are removed from all released text, with each detail an LLM flags for removal reviewed by hand (Appendix~\ref{app:impl}).


\bibliography{custom}

\clearpage
\appendix

\section{Implementation and Release}
\label{app:impl}

\paragraph{Judge details.}
Our main judge reads the posting description and the whole
deliverable. Its three votes agree at Fleiss $\kappa$ 0.86--0.87 on
on-spec, client-plausible and pay-worthy (0.75 on pay-worthy when reading only the opening), and 0.67 on fabricated (every item is produced, so $\kappa$ is undefined there). Pay-worthy and client-plausible agree at $\kappa=0.96$, and their forecast AUCs differ by at most 0.01.

\paragraph{Failure reasons.} Each rejecting vote names one label from the prompt's list, and a rejected item takes the most frequent label among the votes that reject it, with ties going to the label listed first. The six causes group the labels as follows: scoping (\texttt{wrong-scope}), missing information (\texttt{missing-context}), modality or harness (\texttt{no-deliverable}, \texttt{harness-limit}), verification
(\texttt{unverifiable}), execution (\texttt{capability-gap}) and
fabrication (\texttt{fabricated}).

\paragraph{Release.}
The \bench{} release ships:
(i) the 1{,}494 \bench{} items with priced values and strata, every label of \S\ref{sec:robust} and every judge vote behind them; (ii) all prompt templates verbatim: Task~A, the judges, the generic and limits Task~B prompts with every ablation variant, and the tool agent's and simulated client's prompts; (iii) every forecaster's Task~B responses under each prompt, every agent deliverable (the benchmark run, the two repeat attempts, the weaker solver's attempts and the tool agent's files, client exchanges and private facts), redacted as described below, and the per-call token log; (iv) scoring and panel code
with internal endpoints removed; (v) cluster-level demand labels and a coarsened cluster$\times$month dollar panel; and (vi) paraphrased surrogates of all 1{,}494 postings, so the benchmark can be re-run without platform access. Item identifiers are
opaque content hashes.

\paragraph{Redaction.} All released text (surrogates, deliverables and their file names, judge votes, client exchanges, anchor postings and criteria) is redacted. The platform's name is replaced, and we remove contact details, links and credentials from the posting, the names of the client's organization, brand, product or project and of private parties, person names, and client-specific account, case or job numbers. Candidates were extracted per posting by an LLM and reviewed by hand; names of well-known third-party tools, public institutions and places are kept, and a pattern scan for leaks runs over all released text. \emph{Surrogate fidelity:} claude-sonnet-5 compared every surrogate
with its original posting for task semantics, stated requirements, level of detail and identifying content.

\paragraph{Layout and verification.} Each item ships as a folder holding the posting, metadata and the solver's deliverable broken into its original files (\texttt{.docx}, \texttt{.xlsx}, \texttt{.csv}, code, images). The anchor (\S\ref{sec:hapi}) ships in the same layout with a surrogate of the real posting, redacted acceptance criteria, labels and claude-opus-5's attempt; the real freelancer's files are listed by type and size only, and one job without a surrogate is withheld. An automated verifier recomputes the paper's counts, rates and AUCs from the released files alone, except those of the human ratings (Appendix~\ref{app:human}); all checks pass on the current release.
\section{Corpus and Pricing}
\label{app:corpus}

The extract has one row per job post. Fixed-price posts carry a budget; most hourly posts disclose a rate bound. We drop budgets above \$1M or rates above \$5k/hour (test postings; 3 rows). Realized demand counts postings with at least one awarded contract (42.6\%). For an hourly post we impute contract value as the midpoint rate times an expected-hours factor from the duration band. An hourly post without a stated rate carries no value. Dollar totals depend on this imputation, but cluster rankings barely move (Spearman $\rho\geq0.98$).

Beyond Figure~\ref{fig:outcomes}a, we do not discuss dollar-weighted acceptance rates in the main text, because they mostly reflect pay type. Weighted by contract value, 69\% of the priced jobs' dollars go to accepted jobs, against 51\% of priced jobs by count. Hourly jobs, whose values are imputed, hold 81\% of the priced dollars and are accepted more often than fixed-price jobs (Table~\ref{tab:percat}). Among fixed-price jobs, whose budgets the clients set, the dollar-weighted rate is 33\%, below the unweighted 43\%, so the agent does not do better on larger jobs. The 350 hourly jobs without a stated rate are accepted at 67\%.

\begin{table}[t]
\centering\small
\caption{From the corpus to \bench{} (\S\ref{sec:samples}). Each row is the count left after that step.}
\label{tab:funnel}
\setlength{\tabcolsep}{3pt}
\begin{tabularx}{\linewidth}{@{}>{\raggedright\arraybackslash\hangindent=0.8em}X r@{}}
\toprule
Step & Left \\
\midrule
Corpus, 2020 to mid-2026 (2{,}389 daily files) & 18.77M \\
1st and 15th of each month (158 days, 79 months); up to 15 per L3 category (205) per day, in file order; descriptions stored to 2{,}000 characters & 286{,}777 \\
Stratified draw: L1 $\times$ month (948 cells), proportional, $\geq$1 per cell, de-duplicated on title and description & 4{,}000 \\
Description over 150 characters; shuffled once (seed 17) & 3{,}408 \\
Walked in order until 1{,}500 kept & 2{,}614 \\
\quad excluded: fewer than four votes for a defined deliverable & $-$1{,}114 \\
Defined deliverable (at least four of the seven base forecasters' modality annotations) & 1{,}500 \\
Solver returned a deliverable & 1{,}494 \\
\bottomrule
\end{tabularx}
\end{table}

\section{Label Noise and Altered Work}
\label{app:noise}
\begin{figure}[t]
\centering
\includegraphics[width=\linewidth]{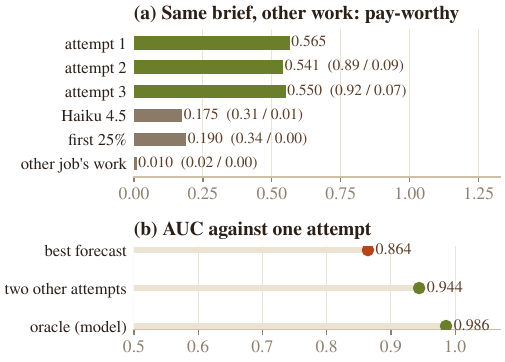}
\caption{Label noise on 400 items (headline judge). (a)~Share pay-worthy with the brief fixed and the work varied; in parentheses, the share among items whose first attempt passes / fails. (b)~AUC against one attempt; the oracle is a beta-binomial predictor knowing each item's true pass rate.}
\label{fig:noise}
\end{figure}
On the 400 items of \S\ref{sec:noise} we check two things (Figure~\ref{fig:noise}): whether the label repeats, and whether the judge grades the work or only the brief.

\textbf{The label repeats.} The agent did each job two more times, and the judge voted again on the same deliverables. Two attempts at the same job agree at Cohen $\kappa=0.81$, and fresh votes on the same deliverable agree at $\kappa=0.87$. About four fifths of the variation in the verdict lies between jobs, not between attempts or votes, so the label is mostly a property of the job. This also caps what any forecast can reach (Figure~\ref{fig:noise}b): the agent's other two attempts predict a third at AUC 0.94, and a predictor that knew each job's true pass rate would reach 0.99. The best forecast reaches 0.86, so it recovers most, but not all, of what can be predicted.

\textbf{The judge reads the work.} Keeping each brief, we replaced the agent's deliverable with work that should fail more often: a weaker solver's attempt (claude-haiku-4.5), the first quarter of the agent's own deliverable, and another job's deliverable from the same category. The agent's own attempts pass 54--57\% of the time; these pass 17.5\%, 19\% and 1\% (Figure~\ref{fig:noise}a). A judge that reacted only to the brief would pass them about as often as the agent's own work.

\section{External Anchor: The Judge on Paid Work}
\label{sec:hapi}
The anchor asks where the judge's bar sits relative to work real clients accepted. We grade 322 curated, human-checked
jobs outside the benchmark where a client paid for a freelancer's
submission (for a few, a submission was also rejected), using the
extracted text and the benchmark prompt, without telling the judge that the
work is human; claude-opus-5 attempts the same jobs with the \bench{} harness.

The judge does not share these clients' bar. Reading in full, it judges
only 16.5\% of work a client already paid for to be pay-worthy, and it
accepts more of claude-opus-5's attempts at the same jobs (27.8\%); reading
only the opening, the two are close (Table~\ref{tab:anchor}). Reading the
opening, the judge passes none of the ten rejected submissions, but with
ten, its separation of paid from rejected work (AUC 0.60 in full, 95\% CI
0.47--0.70) does not exclude 0.5. The judge also sees only the final
submission, not the exchanges and material that preceded it. Every anchor
job was eventually paid, so the anchor cannot score forecasters, and its
jobs are curated, not sampled.

We therefore do not read the gap as the agent's work being better than the freelancers'. A paying client judges a submission with the exchanges and files behind it, which the judge lacks, so the anchor shows only that the judge's acceptance rate is not a real client's. Whether these clients would accept the agent's attempts is left open; answering it would need their own verdicts on that work.

\section{Human Raters}
\label{app:human}
The audit asks how often human raters confirm the judge's accepts and
whether forecasts rank the raters' verdicts. The pool is the 825 items the headline judge called client-plausible or pay-worthy (812 pay-worthy). Each category went to one rater, and two raters with hiring or delivery experience in their categories covered six categories each. Each category's items were shuffled with a fixed seed and raters worked from the top of their file, so the 149 rated items are about twelve per category (7--29); 135 of them form an unbroken block from the top of their file, a random sample of that category's accepts, and the results hold on those
alone. Raters saw the posting, its budget line and the whole deliverable with its files, as the judge saw them, and scored the judge's four criteria with its definitions, fabrication rule and failure labels (with short glosses of the label
names); they were not told the judge's verdict. The headline judge rates 144 of the 149 items pay-worthy, so forecast AUCs
against the raters' verdict measure what the forecasts add beyond the
judge's. As in the main results (\S\ref{sec:ablation}), the limits gain works through the jobs' need for the client's material. Within-category AUCs count only pairs of items from the same category. Table~\ref{tab:human} gives the
results.

\begin{table*}[t]
\centering\small
\caption{Human raters on 149 items the headline judge accepted
(Appendix~\ref{app:human}). Shares are of the rated items unless noted; brackets are 95\% bootstrap CIs (resampling within category for the pay-worthy rate). AUCs are against the raters' pay-worthy call, and ranges are across the eighteen forecasters; $\Delta$ is a paired gain of the limits over the generic prompt.}
\label{tab:human}
\renewcommand{\arraystretch}{1.12}
\newcommand{\grp}[1]{\multicolumn{3}{@{}l}{\textbf{#1}}\\[1pt]}
\begin{tabularx}{\textwidth}{@{}>{\raggedright\arraybackslash\hangindent=0.8em}X r >{\raggedright\arraybackslash}p{6.1cm}@{}}
\toprule
Check & $n$ & Result \\
\midrule
\grp{The raters on the judge's accepts}
Produced / on-spec: raters (judge) & 149 & 99.3 / 82.6\% (100 / 94.0\%) \\
Pay-worthy, raters & 149 & 34.9\% [28.2, 42.3] \\
Raters' rejections labelled missing information & 97 & 75.3\% \\
\midrule
\grp{What the job needs}
Raters' pay-worthy by the needs label: text only / client files & 60 / 53 & 53.3 / 17.0\% \\
Rejected for missing information: shortest / middle / longest third of briefs & 50 / 49 / 50 & 58.0 / 53.1 / 36.0\% \\
Deliverable length & 149 & AUC 0.45 [0.35, 0.55] \\
\midrule
\grp{Forecasts against the raters' verdict}
Generic / limits forecasts & 146--147 & AUC 0.46--0.60, no CI above 0.5 / \newline 0.55--0.70, 15 of 18 CIs above 0.5 \\
Mean over eighteen, items resampled jointly & 145 & generic 0.54 [0.44, 0.63], limits 0.65 [0.57, 0.73]; \newline $\Delta$ $+0.12$ [0.04, 0.20] \\
$\Delta$ per forecaster; AUC within category & 146--147 & $+0.04$ to $+0.19$, 12 of 18 CIs exclude 0; \newline generic 0.45--0.62, limits 0.55--0.72, limits higher for all 18 \\
Means, unbroken top block only (items every forecast covers) & 131 & generic 0.53, limits 0.64 \\
Averaged limits forecast / text-embedding classifier fitted to the judge's verdicts (cross-fitted) & 145 & AUC 0.68 [0.58, 0.77] / 0.53 [0.43, 0.63]; \newline $\Delta$ $+0.15$ [0.03, 0.27] \\
\bottomrule
\end{tabularx}
\end{table*}


\section{Elicitation Detail}
\label{app:mech}

Every Task~B prompt shows the posting as a freelancer
sees it: title, description,
skill tags, budget or
hourly range, engagement type and duration, and expertise tier. The
\emph{generic} prompt then describes ``an autonomous AI agent with
frontier capability for generation, equipped with a browser, code execution,
and file read/write, that could take this job from acceptance to delivery
with no human in the loop besides the client'', under a system message
that calls the agent tool-using. The \emph{limits} prompt keeps the posting
block and the requested fields but replaces the agent description with the
six limits of \S\ref{sec:harness} and uses a system message that refers
only to ``the AI agent described in the request''. The ablation prompts
drop one limit or keep only one; the placebo keeps the frame and list
shape with six neutral statements of similar length (e.g.\ that the agent
reads the whole posting and works on one job at a time). A mis-stated
variant of the limits prompt, which claims a code sandbox the agent lacks
and keeps the tool-using system message, also beats the generic prompt,
but the accurate wording beats it for all seven base forecasters
(Table~\ref{tab:limits}). 

Each forecast is one JSON object: $\hat p$ and the
five annotations of Table~\ref{tab:axes}, on the scales stated in the
prompt (context
dependence 1--5, verifiability 1--5, autonomy horizon 1--5,
tool/credential surface 0--3, deliverable modality 0/1).

Models on the main gateway run at
temperature 0; reasoning models run at
their default effort, without a temperature. Models on
the second route run at low effort. A reply counts only if it parses as
JSON with $\hat p\in[0,1]$; unparsed items are re-queried once with twice
the budget, and items that still fail are left out of that forecaster's
scores, so $n$ varies slightly by forecaster and prompt
(Figure~\ref{fig:forecasters}).

Table~\ref{tab:models} gives every model
by the exact identifier sent to its provider and the dates of our calls.
Only claude-sonnet-4.5 and claude-haiku-4.5 carry a dated snapshot; the
other identifiers are aliases that resolve to the provider's version at
call time, and gpt-6-astra, gpt-5.6-luna and gpt-5.6-sol are the names under
which the second route serves those models.

\begin{table}[h]
\centering\small
\caption{Models by the identifier sent to the provider, their roles
(F forecaster, S solver, J judge, C the tool agent's simulated client, A
annotation of surrogates or reasons, E embeddings) and the first and last day we called them (2026). The text writes version separators as dots and drops snapshot dates: \texttt{glm-5p3} is glm-5.3, \texttt{qwen3p8-max} qwen3.8-max, \texttt{claude-opus-5-5} claude-opus-5.5, and \texttt{claude-sonnet-4-5-20250929} and \texttt{claude-haiku-4-5-20251001} are claude-sonnet-4.5 and claude-haiku-4.5.}
\label{tab:models}
\setlength{\tabcolsep}{3pt}
\begin{tabular}{@{}llr@{}}
\toprule
Identifier & Role & Calls \\
\midrule
\texttt{gpt-5} & F, J & 09-20--10-11 \\
\texttt{gpt-5-mini} & F, C & 09-20--10-11 \\
\texttt{gpt-5-nano} & F & 09-21--10-11 \\
\texttt{gpt-4.1} & F, A & 09-21--10-11 \\
\texttt{gpt-4o} & F & 09-21--10-11 \\
\texttt{o3} & F & 09-21--10-11 \\
\texttt{gpt-5.1} & F & 09-24--10-11 \\
\texttt{o4-mini} & F & 09-24--10-11 \\
\texttt{gpt-6-astra} & F & 09-25--10-11 \\
\texttt{gpt-5.6-luna} & F & 09-25--10-11 \\
\texttt{gpt-5.6-sol} & F & 09-25--10-11 \\
\texttt{claude-sonnet-4-5-20250929} & F & 09-20--10-11 \\
\texttt{claude-opus-5} & F, S & 09-30--10-11 \\
\texttt{claude-opus-5-5} & F & 10-01--10-11 \\
\texttt{claude-fable-5-1} & F, S & 10-01--10-11 \\
\texttt{claude-haiku-4-5-20251001} & S & 10-06--10-08 \\
\texttt{claude-sonnet-5} & J, A & 10-02--10-08 \\
\texttt{glm-5p3} & F, J & 10-01--10-11 \\
\texttt{kimi-k3} & F, J & 10-01--10-11 \\
\texttt{qwen3p8-max} & F, J & 10-01--10-08 \\
\texttt{text-embedding-3-small} & E & 09-20--09-24 \\
\bottomrule
\multicolumn{3}{@{}l}{}
\end{tabular}
\end{table}

\begin{table}[h]
\centering\small
\caption{Annotation axes elicited alongside each forecast (Task B).}
\label{tab:axes}
\begin{tabular}{@{}lL{4.2cm}@{}}
\toprule
Axis & Question put to the model \\
\midrule
Context dependence & Needs the client's private assets, brand voice,
prior work, tacit taste \\
Verifiability & Can success be judged from the posting alone \\
Autonomy horizon & Steps before a human must intervene \\
Tool/credential surface & Needs accounts, payments, physical access \\
Deliverable modality & Emittable artifact vs.\ ongoing service \\
\bottomrule
\end{tabular}
\end{table}

\begin{table}[h]
\centering\small
\caption{Posting features in a standardized logistic regression of the
headline label on the 1{,}474 \bench{} jobs that gpt-5-mini annotated. The fit also
includes gpt-5-mini's five annotation axes (Table~\ref{tab:axes}) as
controls. Log budget is taken over priced jobs; the hourly jobs without a
stated rate get the priced mean and their own indicator.}
\label{tab:mech}
\setlength{\tabcolsep}{4pt}
\begin{tabular}{@{}lrr@{}}
\toprule
Predictor & $\hat\beta$ & $p$ \\
\midrule
Log posted budget       & $-0.25$ & 0.003    \\
Hourly pay              & $+0.60$ & $<$0.001 \\
Hourly, no stated rate  & $-0.14$ & 0.088    \\
Log description length  & $+0.11$ & 0.078    \\
Skill count             & $+0.03$ & 0.570    \\
\bottomrule
\end{tabular}
\end{table}

\section{Other Judges, Solvers and Biases}
\label{app:judges}

\begin{figure}[t]
\centering
\includegraphics[width=\linewidth]{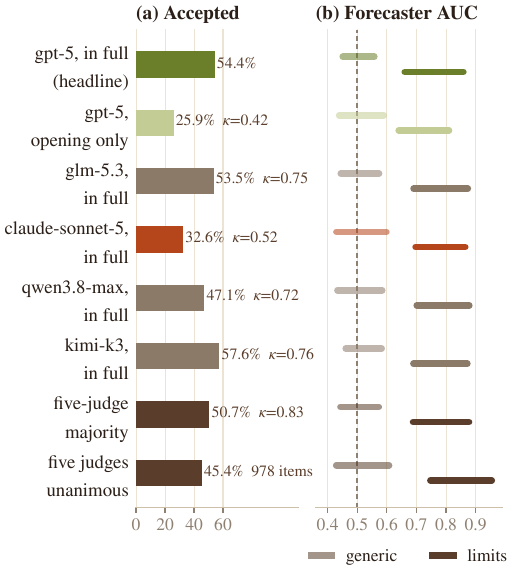}
\caption{The claude-opus-5 deliverables graded by judges from
five frontier model families (three votes each, same prompt; claude-sonnet-5 with thinking
disabled), their majority, and the items on which all five agree.
(a)~Share accepted and Cohen $\kappa$ against the headline label.
(b)~Range of forecaster AUC against that label over the eighteen
forecasters, generic (light) and limits prompt (dark); dashed: random ranking (0.5).}
\label{fig:judges}
\end{figure}

LLM judges can favor outputs from their own model family, and the main judge is an
OpenAI model grading an Anthropic solver. We therefore graded the same deliverables with judges from four other frontier model families, with the same
prompt and three votes each, reading in full as the headline judge does
(Figure~\ref{fig:judges}, Table~\ref{tab:labels}). Pairs of judges agree
at Cohen $\kappa$ 0.49--0.79. claude-sonnet-5 ran
with extended thinking disabled and is the strictest of the five. The panel majority counts
each judge once, by its own three-vote majority, on items graded by at
least four judges (claude-sonnet-5 is missing two).

\paragraph{Other solvers.} On the 400 items of \S\ref{sec:noise},
claude-fable-5-1 and the weaker solver there, claude-haiku-4.5, attempted
every job with the same harness; Fable's verdicts agree with the agent's at
$\kappa=0.75$. Against each solver's outcomes, generic forecasts reach AUC
0.42--0.54 and 0.43--0.61, limits forecasts 0.68--0.87 and 0.68--0.85, and
every forecaster gains, with every 95\% CI above zero. Haiku passes almost
only jobs the benchmark agent passes; within those 226 jobs, limits
forecasts pick out the ones Haiku passes (AUC 0.63--0.74, all CIs above
0.5, against 7 of 18 generic forecasts), so they track a difficulty
gradient the solvers share. Nor does a model know its own work best: with
forecaster and solver effects removed, claude-opus-5 ranks its own attempts
no better than others do ($-0.011$, CI $-0.022$ to $0.001$;
Table~\ref{tab:protocol}; cf.\ \citealp{barkan2025capable}).

\paragraph{Own-family bias.} We double-center the 18$\times$5 table of
forecaster-by-judge AUCs and compare the mean residual of the 18
same-family pairs with that of the others (item bootstrap, 500 resamples).
With forecaster and judge effects removed, forecasters rank the judge from
their own model family better by $+0.003$ AUC (95\% CI 0.001--0.005) with
the limits prompt and $+0.007$ (0.004--0.009) with the generic one, against
limits gains of $+0.20$ to $+0.40$.

\paragraph{Deliverable length and checkable jobs.} The judge favors longer
deliverables \citep[cf.][]{dubois2024length}: length alone ranks the
headline label at AUC 0.64, a preference human raters do not share (AUC
0.45 on the judge's accepts; Table~\ref{tab:human}), but limits forecasts
hardly track length and hold their AUC within each fifth of the
deliverable-length distribution. On the 449 jobs rated clearly verifiable
(median verifiability $\geq4$ of 5), generic forecasts stay near random and
limits forecasts reach 0.66--0.90. With fabrication-flagged deliverables
dropped and on the paraphrased surrogates, the results hold under both
prompts (limits AUC 0.67--0.86; Table~\ref{tab:selection}). Rewrites that
leave a task's requirements unchanged can still change how models perform
on it \citep{rezaei2026frail}, but
paraphrasing moves no base forecaster's AUC by more than 0.015.

\section{The Agent the Generic Prompt Describes}
\label{sec:nearchance}
Generic forecasts might fail only because they describe a different agent from ours, one with tools and a client in the loop. To test this, we ran such an agent on 500 jobs: claude-opus-5 in a multi-turn loop with web search, a code sandbox and file read/write, able to send up to five messages to a simulated client. The simulated client holds private facts about the job, which it generated beforehand, and can paste text; the judge also sees these facts and the exchange. The agent has every tool the generic prompt names and keeps only three limits that prompt does not mention: no access to the client's systems, which real postings rule out, and named text-encodable files (L3--L5; \S\ref{sec:harness}).

\begin{figure}[t]
\centering
\includegraphics[width=\linewidth]{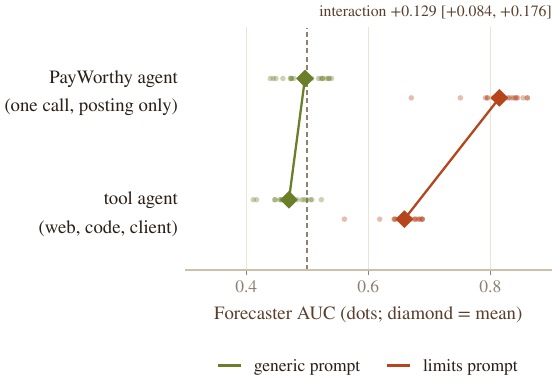}
\caption{Generic and limits forecasts against the outcomes of two agents on the same 500 jobs: the \bench{} agent, and the tool-using agent with a simulated client that the generic prompt
describes. Dots: one forecaster each
(18 per prompt); diamonds: mean AUC. The two agents' verdicts agree
at Cohen $\kappa=0.29$. The interaction is the generic prompt's advantage on the tool agent minus
the limits prompt's, with a 95\% bootstrap CI over jobs.}
\label{fig:toolagent}
\end{figure}

The judge accepts this tool agent's work more often than our agent's (67\% against 52\% on the same jobs), yet generic forecasts rank its jobs near random too (mean AUC 0.47; Figure~\ref{fig:toolagent}). Limits forecasts, written for our agent, rank its jobs better (AUC 0.56--0.69), possibly through the shared limits, though switching agents costs them 0.13 more AUC than the generic ones. So generic forecasts fail even for the agent they describe.

\section{Additional Results}
\label{app:extra}

\begin{figure}[t]
\centering
\includegraphics[width=\linewidth]{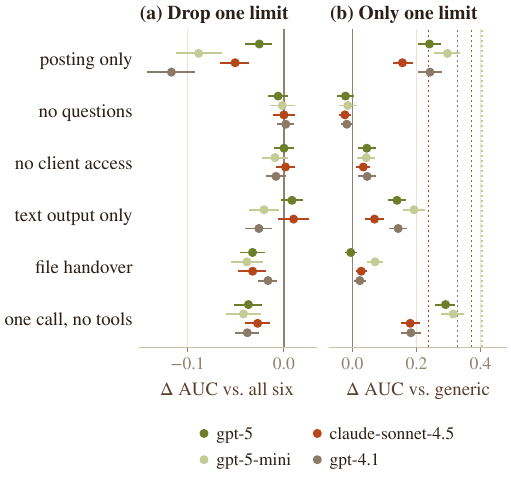}
\caption{Ablating the limits prompt (four forecasters, headline label,
95\% paired-bootstrap CIs). (a)~AUC change when one limit is dropped from
the full prompt. (b)~AUC gain over the generic prompt when only one limit
is stated; dashed lines mark each forecaster's full-prompt gain.}
\label{fig:ablation}
\end{figure}

\begin{table}[t]
\centering\small
\caption{External anchor and reading length: the \bench{} judge (gpt-5,
three votes) on real submissions that a client paid for and on
claude-opus-5's attempts at the same jobs, reading only the opening or the
whole deliverable. Share judged pay-worthy unless noted; brackets are 95\%
bootstrap CIs.}
\label{tab:anchor}
\renewcommand{\arraystretch}{1.12}
\newcommand{\grp}[1]{\multicolumn{3}{@{}l}{\textbf{#1}}\\[1pt]}
\begin{tabularx}{\linewidth}{@{}>{\raggedright\arraybackslash\hangindent=0.8em}X r >{\raggedright\arraybackslash}p{2.75cm}@{}}
\toprule
Check & $n$ & Result \\
\midrule
\grp{Anchor: real submissions to 322 jobs}
Paid work, opening / full & 256 / 255 & 19.9\% [15.2, 24.6] \newline 16.5\% [12.2, 20.8] \\
claude-opus-5, opening / full & 321 / 320 & 19.0\% [14.6, 23.4] \newline 27.8\% [22.8, 32.8] \\
AUC, paid vs.\ the 10 rejected submissions, opening / full & 256 / 255 & 0.63 [0.53, 0.69] \newline 0.60 [0.47, 0.70] \\
\midrule
\grp{Reading length: \bench{}}
Pay-worthy rate, full (headline) & 1{,}494 & 54.4\% [51.9, 57.0] \\
Pay-worthy rate, opening & 1{,}494 & 25.9\% [23.7, 28.0] \\
Deliverables longer than the opening window / still cut at the output limit & 1{,}494 & 99.7\% / 0.6\% \\
\bottomrule
\end{tabularx}
\end{table}

\begin{table}[t]
\centering\small
\caption{Forecasts against eight labels of the same deliverables:
gpt-5 reading the whole deliverable (headline) or the opening only, four
judges from other model families reading in full, the majority of the five
full-reading judges, and the 978 items on which all five agree. Generic
and limits: AUC range over the eighteen forecasters; gain: range of paired
gains of the limits over the generic prompt (every 95\% CI excludes 0;
lowest lower bound $+0.147$). Text: 5-fold cross-validated logistic regression on
text-embedding-3-small embeddings of the title, description and skills. $\kappa$ is against the headline label (on the unanimous
items it is 1 by construction).}
\label{tab:labels}
\setlength{\tabcolsep}{2.5pt}
\begin{tabular}{@{}lrrcccc@{}}
\toprule
Label & Rate & $\kappa$ & Generic & Limits & Gain & Text \\
\midrule
gpt-5 (headline) & 54.4 & -- & .45--.56 & .66--.86 & .20--.40 & .82 \\
gpt-5, opening & 25.9 & .42 & .44--.59 & .64--.81 & .18--.33 & .73 \\
glm-5.3 & 53.5 & .75 & .44--.57 & .69--.87 & .25--.42 & .81 \\
claude-sonnet-5 & 32.6 & .52 & .43--.60 & .70--.86 & .22--.40 & .80 \\
qwen3.8-max & 47.1 & .72 & .43--.58 & .70--.88 & .25--.42 & .84 \\
kimi-k3 & 57.6 & .76 & .46--.58 & .69--.87 & .23--.41 & .81 \\
\midrule
Panel majority & 50.7 & .83 & .44--.57 & .69--.88 & .25--.43 & .83 \\
Panel unanimous & 45.4 & -- & .43--.61 & .75--.95 & .29--.50 & .92 \\
\bottomrule
\end{tabular}
\end{table}

\paragraph{Annotations versus probabilities.} Refit on outcomes, the five generic-prompt annotations rank the jobs better
than the stated probabilities: they reach AUC 0.70 for gpt-5-mini, whose
probability ranks below random, and for the other base forecasters they beat
noise features and match or exceed random projections of the description
embedding (Table~\ref{tab:selection}). The annotations carry signal that the probabilities leave unused, though a
classifier fitted on outcomes learns the judge's preferences, whatever the
model itself understands.

\paragraph{Fabrication.} Reading in full, the judge's majority flags four
deliverables (0.3\%), all rejected; reading only the opening, it flags
6.4\%. Dropping the four leaves every AUC range unchanged.

\paragraph{Checks.} Table~\ref{tab:selection} collects the checks on
selection, posting features, annotation refits and surrogates, and
Tables~\ref{tab:needs}--\ref{tab:protocol} those on missing information,
what the forecasts track, the limits prompt and the protocol. Every row is computed from released files.

\begin{table*}[t]
\centering\small
\caption{Checks on selection, posting features, annotation refits and
surrogates, on the 1{,}494 \bench{} items (headline label). AUC ranges are
across forecasters.}
\label{tab:selection}
\renewcommand{\arraystretch}{1.12}
\newcommand{\grp}[1]{\multicolumn{3}{@{}l}{\textbf{#1}}\\[1pt]}
\begin{tabularx}{\textwidth}{@{}>{\raggedright\arraybackslash\hangindent=0.8em}X r >{\raggedright\arraybackslash}p{6.1cm}@{}}
\toprule
Check & $n$ & Result \\
\midrule
\grp{Selection and posting features}
Own annotation left out of the \bench{} selection (base seven) & 1{,}384--1{,}425 & AUC 0.456--0.557 (full rule 0.456--0.558) \\
Reweighted to corpus shares of postings / dollars: by category; by cluster & 1{,}494 & 54.4\% / 57.6\%; 51.0\% / 56.3\% (as sampled 54.4\%) \\
Spearman of contract value with outcome, priced jobs: fixed-price / hourly & 754 / 390 & $-0.05$ / $+0.20$ \\
Spearman with outcome: log description length / skill count & 1{,}494 / 1{,}429 & $-0.01$ / $+0.08$ \\
\midrule
\grp{Annotations refit on outcomes (cross-fitted)}
Five generic-prompt annotation axes, gpt-5-mini / other six base forecasters & 879--1{,}474 & AUC 0.703 / 0.585--0.665 \\
Five Gaussian noise features, 20 draws & 1{,}494 & AUC 0.494 (0.443--0.522) \\
Five random projections of the description embedding, 20 draws & 1{,}494 & AUC 0.585 (0.540--0.652) \\
\midrule
\grp{Surrogates}
Description length, surrogate vs.\ original (Spearman; median length ratio) & 1{,}489 & 0.96 (1.00) \\
Generic forecast AUC on surrogate vs.\ original text (base seven) & 1{,}469--1{,}474 & AUC 0.457--0.542 (orig.\ 0.456--0.558); \newline all $|\Delta|\le0.015$ \\
Limits forecast AUC on surrogate vs.\ original text (base seven) & 1{,}475--1{,}485 & AUC 0.672--0.863 (orig.\ 0.661--0.858); \newline all $|\Delta|\le0.014$ \\
\bottomrule
\end{tabularx}
\end{table*}

\begin{table*}[t]
\centering\small
\caption{What the posting needs and missing information, on the 1{,}494
\bench{} items (headline label). The AUC range is across forecasters;
``mean'' is the mean generic forecast of the fifteen OpenAI and Anthropic
forecasters; brackets are 95\% bootstrap CIs.}
\label{tab:needs}
\renewcommand{\arraystretch}{1.12}
\begin{tabularx}{\textwidth}{@{}>{\raggedright\arraybackslash\hangindent=0.8em}X r >{\raggedright\arraybackslash}p{6.1cm}@{}}
\toprule
Check & $n$ & Result \\
\midrule
Generic forecasts with missing-information rejections set aside & 1{,}223 & AUC 0.521--0.630; mean 0.584 [0.547, 0.618] \\
Missing-information rejection rate, short / long third of briefs & 1{,}489 & 20.9\% / 15.1\% \\
The needs label, annotated by claude-sonnet from the posting: text only / client files / client-system access / a medium the agent cannot produce & 1{,}494 & share 25 / 43 / 18 / 14\%; \newline pay-worthy 82.3 / 41.8 / 63.7 / 32.1\% \\
\bottomrule
\end{tabularx}
\end{table*}

\begin{table*}[t]
\centering\small
\caption{What the forecasts track, on the \bench{} items (headline
label; eighteen forecasters unless noted). AUC ranges are across
forecasters.}
\label{tab:track}
\renewcommand{\arraystretch}{1.12}
\begin{tabularx}{\textwidth}{@{}>{\raggedright\arraybackslash\hangindent=0.8em}X r >{\raggedright\arraybackslash}p{6.1cm}@{}}
\toprule
Check & $n$ & Result \\
\midrule
Separating self-contained jobs from jobs needing the client's files: generic / limits forecasts (outcome 0.70) & 999--1{,}016 & AUC 0.41--0.57, 7 of 18 CIs below 0.5 / 0.69--0.87, all CIs above 0.5 \\
Top fifth by each generic forecast (ties broken at random, mean of 20 draws): share needing the client's files / a medium the agent lacks (all jobs 43 / 14\%); acceptance of its client-file picks (all client-file jobs 42\%); share of the top fifth that the same forecaster's limits forecast puts in its top fifth / bottom half & 1{,}472--1{,}494 & 51--62\% (median 58) / 0--5\% (median 3); \newline 26--40\% (median 31); \newline 23--53\% (median 38) / 21--54\% (median 37) \\
Forecast (logit) explained by posting size (budget, description length, skills, pay type, tier) / needs label / own five annotations, cross-validated $R^2$, medians & 1{,}472--1{,}494 & generic $\leq$0.03 (max) / 0.21 / 0.45; \newline limits $\leq$0.01 (max) / 0.34 / 0.25 \\
One-line reasons (gpt-5, gpt-5-mini, claude-sonnet-4.5, claude-opus-5; 300 items of the 400; cues tagged by gpt-4.1): share naming missing client material as a risk; mean forecast when they do / do not & 1{,}199 per prompt & generic 37\%: 0.50 / 0.49; \newline limits 61\%: 0.15 / 0.20 \\
The same four with the reason field: AUC (without the field) & 299--300 & generic 0.44--0.59 (0.44--0.54); \newline limits 0.79--0.88 (0.80--0.88) \\
\bottomrule
\end{tabularx}
\end{table*}

\begin{table*}[t]
\centering\small
\caption{Checks on the limits prompt, on the 1{,}494 \bench{} items
(headline label; eighteen forecasters unless noted). AUC ranges are across
forecasters; brackets are 95\% bootstrap CIs; $\Delta$ is a paired gain of
the limits over the generic prompt.}
\label{tab:limits}
\renewcommand{\arraystretch}{1.12}
\begin{tabularx}{\textwidth}{@{}>{\raggedright\arraybackslash\hangindent=0.8em}X r >{\raggedright\arraybackslash}p{6.1cm}@{}}
\toprule
Check & $n$ & Result \\
\midrule
The needs label alone, cross-fitted & 1{,}494 & AUC 0.69; below 17 of 18 limits forecasts \\
Hand-written rule on the needs label, no labeled jobs: low if the job needs the client's files / low unless self-contained & 1{,}494 & AUC 0.61 [0.58, 0.63] / 0.64 [0.62, 0.66] \\
Forecast minus the rule (low unless self-contained), paired: generic / limits & 1{,}472--1{,}494 & $-0.19$ to $-0.08$, all CIs below 0 / \newline $+0.02$ to $+0.22$, 17 of 18 CIs above 0 (client-files rule: $+0.05$ to $+0.25$, all CIs above 0) \\
AUC within jobs of the same needs label: generic / limits & 1{,}472--1{,}494 & 0.39--0.49, 10 of 18 CIs below 0.5 / \newline 0.60--0.83, all CIs above 0.5 \\
$\Delta$ within jobs needing the client's files / system access & 633 / 256 & $+0.21$ to $+0.46$ / $+0.22$ to $+0.36$; all CIs exclude 0 \\
$\Delta$ within self-contained jobs / jobs needing a medium the agent lacks & 366 / 215 & $+0.10$ to $+0.36$, all CIs exclude 0 / $-0.00$ to $+0.23$, 12 of 18 CIs exclude 0 \\
$\Delta$ with missing-information rejections set aside & 1{,}220 & $+0.17$ to $+0.33$, all CIs exclude 0 \\
Added to keywords, the needs label and the generic forecast (cross-fitted) & 1{,}465--1{,}485 & AUC $+0.00$ to $+0.13$ (base 0.72--0.74); 17 of 18 CIs exclude 0 \\
Best five limits forecasts minus the text-embedding classifier (AUC 0.82), paired & 1{,}469--1{,}471 & $+0.038$, $+0.037$, $+0.026$, CIs exclude 0; \newline $+0.014$, $+0.012$, CIs include 0 \\
Stacked on the text-embedding classifier (cross-fitted): generic / limits forecast & 1{,}460--1{,}474 & AUC $+0.00$ to $+0.00$, no CI excludes 0 / $+0.01$ to $+0.07$, all 18 CIs exclude 0 \\
Mis-stated limits (sandbox claimed, tool-using system message), base seven & 1{,}488--1{,}494 & AUC 0.538--0.776; accurate wording better by $+0.07$ to $+0.23$, all CIs exclude 0 \\
\bottomrule
\end{tabularx}
\end{table*}

\begin{table*}[t]
\centering\small
\caption{Checks on the protocol and on dependence, on the \bench{} items
(headline label; eighteen forecasters). AUC ranges are across forecasters;
brackets are 95\% bootstrap CIs; $\Delta$ is a paired gain of the limits
over the generic prompt.}
\label{tab:protocol}
\renewcommand{\arraystretch}{1.12}
\begin{tabularx}{\textwidth}{@{}>{\raggedright\arraybackslash\hangindent=0.8em}X r >{\raggedright\arraybackslash}p{6.1cm}@{}}
\toprule
Check & $n$ & Result \\
\midrule
Only deliverables written in a single call & 1{,}158 & generic AUC 0.47--0.56, limits 0.66--0.87; \newline $\Delta$ $+0.18$ to $+0.40$, all CIs exclude 0 \\
Haiku 4.5 passes, within the jobs the benchmark agent passes (400-item subset): limits / generic & 223--226 & AUC 0.63--0.74, all CIs above 0.5 / 0.47--0.63, 7 of 18 \\
claude-opus-5 on its own attempts vs.\ Fable's and Haiku's, double-centred AUC residual: limits / generic & 395--398 & $-0.011$ [$-0.022$, 0.001] / $-0.020$ [$-0.034$, $-0.006$] \\
Mean of two attempts predicting the third: observed / implied by the beta-binomial fit; its oracle AUC (item bootstrap) & 399 & 0.944 / 0.945; 0.99 [0.98, 0.99] \\
Mean over all eighteen, items resampled jointly (items every forecast covers) & 1{,}449 & generic AUC 0.51, limits 0.80; $\Delta$ $+0.30$ [0.27, 0.32]; \newline mean $\Delta$ by model family $+0.26$ to $+0.31$ \\
Limits forecast recalibrated (Platt) on 100 random labeled jobs, scored on the rest (20 splits) & 100 / rest & ECE 0.05--0.11 (uncalibrated 0.22--0.47); \newline accuracy at 0.5: 0.63--0.78 (always accept 0.54) \\
\bottomrule
\end{tabularx}
\end{table*}

\begin{table*}[t]
\centering\small
\caption{Settings at a glance: which agent, forecast prompt and label each
result uses. ``Limits'' and ``generic'' are the two Task~B prompts
(\S\ref{sec:harness}, Appendix~\ref{app:mech}); AUC ranges are across
forecasters.}
\label{tab:settings}
\renewcommand{\arraystretch}{1.12}
\setlength{\tabcolsep}{4pt}
\begin{tabularx}{\textwidth}{@{}>{\raggedright\arraybackslash}p{1.9cm} >{\raggedright\arraybackslash}X >{\raggedright\arraybackslash}p{2.5cm} >{\raggedright\arraybackslash}p{2.9cm} r >{\raggedright\arraybackslash}p{3.2cm}@{}}
\toprule
Setting & Agent & Forecasts & Label & Items & Main result \\
\midrule
Benchmark (\S\ref{sec:harness}) & claude-opus-5; posting only, one response, named files & generic, limits (18 forecasters) & gpt-5, full reading, 3 votes (headline) & 1{,}494 & generic 0.45--0.56; limits 0.66--0.86 \\
Ablation (\S\ref{sec:ablation}) & same & limits with one limit dropped or alone; placebo (4 forecasters) & headline & 1{,}494 & L6 and L1 carry most of the gain \\
Tool agent (App.~\ref{sec:nearchance}) & claude-opus-5 with tools and a simulated client & generic, limits & gpt-5, also sees the client's facts and exchange & 500 & generic near random; limits 0.56--0.69 \\
Label noise (\S\ref{sec:noise}) & benchmark agent, two repeat attempts; a weaker solver; altered work & best limits & headline judge, fresh votes & 400 & attempts $\kappa$ 0.81, votes 0.87; ceiling 0.94 (model-based 0.99) vs.\ 0.86 \\
Other labels (\S\ref{sec:robust}) & benchmark agent & generic, limits & gpt-5 opening only; four other judges; their majority; unanimous items & 1{,}494 (978) & generic 0.43--0.61; limits 0.64--0.95 \\
Other solvers (\S\ref{sec:robust}) & claude-fable-5-1; claude-haiku-4.5; same harness & generic, limits & headline judge & 400 & generic 0.42--0.61; limits 0.68--0.87 \\
Anchor (App.~\ref{sec:hapi}) & claude-opus-5; work real clients paid for & none & headline judge & 322 & paid work passes 16.5\% \\
Human raters (\S\ref{sec:robust}, App.~\ref{app:human}) & benchmark agent; the judge's accepts only & generic, limits & two human raters, the judge's criteria & 149 & raters accept 35\%; generic 0.46--0.60, limits 0.55--0.70 \\
\bottomrule
\end{tabularx}
\end{table*}

\end{document}